\documentclass[11pt,a4paper]{article}
\usepackage[hyperref]{emnlp2020}
\usepackage{graphicx}
\graphicspath{ {./images/} }
\usepackage{times}
\usepackage{latexsym}
\usepackage{url}

\usepackage{microtype}

\aclfinalcopy 

\definecolor{Fern Green}{RGB}{57, 150, 33}
\definecolor{Red}{RGB}{100, 0, 0}
\definecolor{Royal Purple}{rgb}{0.47, 0.32, 0.66}

\title{Data Troubles in Sentence Level Confidence Estimation for Machine Translation}

\author{Ciprian Chelba\Thanks{Corresponding author: ciprianchelba@google.com.} \\
  Google, Inc. \\
  Mountain View, CA, USA \\\And
  Junpei Zhou \\
  LTI, Carnegie Mellon\\
  Pittsburgh, PA, USA \\\And
  Yuezhang (Music) Li \\
  Google, Inc. \\
  Kirkland, WA, USA \\\AND
  Hideto Kazawa \\
  Google, Inc. \\
  Tokyo, Japan \\\And
  Jeff Klingner \\
  Google, Inc. \\
  Mountain View, CA, USA \\\And
  Mengmeng Niu \\
  Google, Inc. \\
  Mountain View, CA, USA \\
  }

\date{}

\begin{document}
\maketitle
\begin{abstract}

The paper investigates the feasibility of confidence estimation for neural machine translation models operating at the high end of the performance spectrum. As a side product of the data annotation process necessary for building such models we propose sentence level accuracy $SACC$ as a simple, self-explanatory evaluation metric for quality of translation. 

Experiments on two different annotator pools, one comprised of non-expert (crowd-sourced) and one of expert (professional) translators show that $SACC$ can vary greatly depending on the translation proficiency of the annotators, despite the fact that both pools are about equally reliable according to Krippendorff's alpha metric; the relatively low values of inter-annotator agreement confirm the expectation that sentence-level binary labeling $good$ / $needs\ work$ for translation out of context is very hard.

For an English-Spanish translation model operating at $SACC = 0.89$ according to a non-expert annotator pool we can derive a confidence estimate that labels 0.5-0.6 of the $good$ translations in an ``in-domain" test set with 0.95 Precision. Switching to an expert annotator pool decreases $SACC$ dramatically: $0.61$ for English-Spanish, measured on the exact same data as above. This forces us to lower the CE model operating point to 0.9 Precision while labeling correctly about 0.20-0.25 of the $good$ translations in the data.

We find surprising the extent to which CE depends on the level of proficiency of the annotator pool used for labeling the data. This leads to an important recommendation we wish to make when tackling CE modeling in practice: it is critical to match the end-user expectation for translation quality in the desired domain with the demands of annotators assigning binary quality labels to CE training data.

\end{abstract}

\section{Introduction}

The quality of machine translation models has improved significantly in recent years to the point that ``in-domain" performance for languages rich in training data has approached ``human parity" \citep{barrault-etal-2019-findings}. Outside this strict context the claim is disputable however: the boundaries of ``in-domain" data are unclear even when the training data that was used to build the model is made available, not to mention the situation of a casual user interacting with the translation model. Idiosyncrasies in the input text can trigger surprising model behaviors even for language pairs at the high end of the translation quality spectrum, \citep{DBLP:journals/corr/abs-1906-11943}. \citet{L_ubli_2020} propose revising the evaluation of strong machine translation systems.

This brings to fore the issue of trustworthiness of a translation service or model, as with the output of pretty much any machine learning algorithm \citep{Spiegelhalter2020Should}\footnote{Also see the NEURIPS 2018 invited talk: ``Making Algorithms Trustworthy: What Can Statistical Science Contribute to Transparency, Explanation and Validation?" \url{https://nips.cc/Conferences/2018/Schedule?showEvent=12346}}: it needs to be accompanied by a \emph{confidence score} to make it truly meaningful and useful.

Besides monolingual users of machine translation (MT) technology (particularly when not knowing the target language), algorithmic uses of MT could also greatly benefit from a sentence-level confidence score. As an example, consider the possibility of translating English web pages and serving them as search results in a different language: keyword hits, titles, snippets should be weighed accordingly when ranking or rendering such content. Another use case is reducing post-editing costs: out of one thousand translations, which is the half or third that is probably correct?

We therefore focus on confidence estimation (CE) for MT: we wish to build a model that can predict whether a given translation is $good$ as is or not. In contrast with the quality evaluation (QE) task \citep{specia2018machine}, we model translation quality as binary valued; to further depart from the QE setup we assume access to both the MT model that produces the translation and to the training data used to build it. We contrast our approach with traditional QE in more detail in Section~\ref{sec:related_work_qe}.

A necessary ingredient is the capability of annotating sentence pairs consisting of source sentence and translation with binary labels $good$ / $needs\ work$. As a side product, this enables a very straightforward and interpret-able way of evaluating the quality of MT models by measuring sentence-level accuracy (SACC). 

We use two different annotator pools for annotation:
\begin{enumerate}
    \item non-experts: proficient in the source and native in the target language
    \item experts: professional translators
\end{enumerate}

With the non-expert pool, in agreement with~\citep{barrault-etal-2019-findings} we find that for high-quality language pairs such as English-German, English-French or English-Spanish, SACC values on ``in-domain" data vary between 0.89 and 0.98; for a lower-quality model such as English-Romanian, SACC is 0.61. A simple model for the reliability of binary annotations allows us to estimate the probability of a annotator assigning the incorrect label on a sample (about 0.05--0.10 in our experiments) and leads to the use of label smoothing in training our CE models.

Switching to the expert pool lowers SACC dramatically: for English-German it drops from 0.98 to 0.57; for English-Spanish from 0.89 to 0.61; for English-French from 0.89 to 0.44 and for English-Romanian from 0.61 to 0.26. The simple Bernoulli annotation model breaks down. 

Analyzing the two annotator pools using Krippendorff's alpha (KA) statistic~\citep{KrippendorffAlpha} for inter-annotator agreement shows that they are both about equally reliable, with the non-expert pool being simply more lenient than the expert one. The low values of KA confirm the expectation that binary translation quality annotation for isolated sentences out of context is very hard.

Practical use cases of MT output are well aligned with CE models in that translation quality is viewed as a binary variable: correct or incorrect, as described in \citep{zhou2020practical}. As a result we suggest measuring CE performance using Recall at a fixed Precision value which is chosen depending on the SACC of the model, e.g. 0.9 or 0.95.

The \emph{utility} of a CE model is however measured by the product of Recall and SACC of the underlying translation model. Indeed, we find that for an English-Spanish MT model whose ``in-domain" performance is 89\% SACC, the CE model can achieve 0.6 Recall at 0.95 Precision, meaning that it can annotate roughly half of the input sentences as $good$ translations with relatively high precision.  

As a more general remark, it is interesting to examine the benefit of CE at various translation quality levels as measured by SACC: 
\begin{itemize} 
\item at the low end of the SACC range, CE at the Precision/Recall operating point mentioned above does not add significant value simply because the ratio of $good$ samples in the data is too low;
\item as SACC increases, e.g. 60\%-90\%, the missed opportunity from not leveraging CE becomes significant: a large percentage of sentences are translated correctly ($good$) and yet we do not know which ones. CE can add significant value in this regime, its utility increasing with the accuracy of the underlying MT model; 
\item finally, when SACC exceeds the desired Precision level for CE (e.g 95\%) we can simply label all translations as $good$, making CE modeling irrelevant.
\end{itemize}

The remainder of the paper is organized as follows: in Section~\ref{sec:problem_def_related_work} we formulate the CE problem for MT and contrast our approach to QE, and relate it to previous work that tackled QE with a binary quality metric. While our approach pertains to the more general body of work on ``uncertainty estimation" for machine learning models, and in particular for seq2seq models~\citep{sutskever2014sequence}, we find little related work in the context of neural MT, as highlighted in~\ref{sec:related_work_ce}. Section~\ref{sec:data} described the data annotation framework and an analysis of the annotator reliability for both non-expert and expert pools. Section~\ref{sec:modeling} describes our modeling approach as an offshoot of standard QE work re-purposed for binary classification augmented with features that are expected to help CE: Monte-Carlo dropout (Section~\ref{sec:modeling:mcdropout}) aiming at model uncertainty and language modeling on source sentences (Section~\ref{sec:modeling:lm}) aiming at mitigating MT behavior on ``out-of-distribution" data. Section~\ref{sec:exps} presents experiments on two data sets. They are derived from the same source and translation pairs by asking two different annotator pools to provide binary labels $good$/$needs work$. The source sentences are ``in-domain" relative to the translation model, sampled from web pages whose translation is requested in a given locale; the ``in-domain"-ness of the data is confirmed by the fact that a language model trained on the source side of the MT training data does not benefit from incremental training on the CE training data.
We follow with conclusions and future work directions in Section~\ref{sec:conclusions}.

\subsection{Contributions}
Besides motivating, developing and evaluating CE models for MT, another contribution is the use of SACC as a metric for MT quality, as described in Section~\ref{sec:data}. 

A second contribution is the use of two annotator pools (one non-expert and one expert) for deriving the labeled data needed for CE. Analyzing the two annotator pools using Krippendorff's alpha (KA) statistic for inter-annotator agreement shows that they are both about equally reliable, with the non-expert pool being simply more lenient than the expert one in terms of SACC. The low values of KA confirm the expectation that binary translation quality annotation for isolated sentences out of context is very hard.

Along with this we develop a quasi-bootstrapping procedure for assigning confidence intervals to SACC measurements and propose a simple probability model for such annotations that we show to be quite accurate for MT models with high SACC according to the non-expert annotator pool. The model allows us to estimate the probability of annotation error and characterize human annotator performance in terms of Precision/Recall. Comparing it to the Precision/Recall attained by our CE models we find that there is significant room for improvement. Alas, the model breaks down for the annotations produced by the expert annotator pool, leaving ample room for research on this topic.

We find surprising the extent to which CE depends on the level of proficiency of the annotator pool used for labeling the data. This leads to an important recommendation we wish to make when tackling CE modeling in practice: matching the end-user expectation for translation quality in the desired domain with the demands of annotators assigning binary quality labels to training data is critical.

\section{Problem Definition and Related Work}
\label{sec:problem_def_related_work}

We take the view that translation quality is binary: the translation output by the MT model for a given input sentence is annotated with either $good$ or $needs\ work$ labels. The CE model is a binary classifier predicting the label. We interpret the probability output by the binary classifier as a measure of trust that the output is $good$, and not one of quality, e.g. $P(good) = 0.5$ does not say anything about the quality of the translation, just that we do not know whether it is $good$ or not.

As a departure from standard QE approaches \citep{specia2018machine}, we assume that a CE model has access to both the MT model producing the translation for a given sentence (``glass-box"), and the training data used for building it. This allows us to factor in both ``model-uncertainty" and ``out-of-distribution" signals when computing the confidence score.

\subsection{Comparison with Sentence Level Quality Estimation in the WMT setup}
\label{sec:related_work_qe}

In addition to the fact that we use a binary cost function for MT quality, there are a few other important differences in our approach relative to the WMT QE task.

Unlike WMT QE, we allow the CE model access to both the model producing the translation and the training data used for building it. In this ``glass-box" approach\footnote{The latest WMT QE eval \url{http://www.statmt.org/wmt20/quality-estimation-task.html} introduces a new Sentence-Level Direct Assessment task that makes available the models used for translation.} the model can leverage encoder/decoder output, or attention matrix values as features that could be relevant to building a better probability model for CE. It can also leverage a language model (LM) on the source side or perhaps a contrastive LM built from the MT model and/or the CE training data to provide a signal describing whether the input sentence is ``in-domain" or not.

In the WMT task definition the QE model is expected to predict the HTER score \citep{snover2006study} for a pair of (source, target) sentences, without access to the MT model that produced the target translation, or knowledge about the training data used. Besides the fact that regression on the HTER score cannot be easily converted to $good / needs\ work$ labels with reasonable accuracy as noted by \citet{turchi2014springer}, this makes the problem unnecessarily hard and somewhat removed from the reality of many MT use cases encountered in practice.

A binary classification view on QE has been advocated before, e.g. \citep{turchi-etal-2013-coping} and \citep{turchi2014springer}. We take the extra step to interpret and evaluate it as a CE problem, annotating data specifically for this purpose.

Our baseline approach re-purposes regression models developed for the WMT QE task, such as \citep{wang-EtAl:2018:WMT4} and we find the data released with the task useful for bench-marking our models. Initial models were developed on the WMT QE data by labeling $good$ = (HTER == 0) and $needs\ work$ otherwise. Unfortunately, due to the fact that the $good$ data has been down-sampled we cannot easily evaluate the SACC of the underlying MT model.

Our approach is better aligned with the very recent work of~\citet{fomicheva2020unsupervised}: besides taking the same ``glass-box" approach we leverage the MT model training data to place a given input sentence in the ``in-domain" to ``out-of-domain" spectrum. Unlike the use of direct assessments (DA) we find the use of a binary quality assessment better suited to practical setups, as explained; moreover, this enables a simple probability model for the data annotation process, estimating the annotator error and deriving a human performance reference point for CE models.

\subsection{Related Work on Confidence Estimation}
\label{sec:related_work_ce}

CE for machine translation is a well established task, however there has not been a lot of work in the context of neural seq2seq models for machine translation.

Our general approach is well aligned with the work of \citep{blatz-etal-2004-confidence}. In our work the underlying MT model is neural and the translation quality is probably higher but the general conclusions regarding CE for MT are still valid. Notably the ``glass-box" view on the MT model is an important departure from the WMT QE setup and quite useful for CE.

\citep{dong-etal-2018-confidence} investigate confidence scoring in the context of seq2seq models \citep{sutskever2014sequence} used for parsing and is the closest to our modeling approach, including Monte-Carlo dropout features and the use of a LM on the input. Our use of encoder embeddings and contrastive LM on the input goes a bit further in deriving features that attempt to model the degree to which the source sentence is ``in-domain". While we use target token level entropy as a feature, sequence level posteriors and entropy is something our model is lacking and an item for future work.

\citep{DBLP:journals/corr/abs-1710-03743} investigates the extent to which the attention matrix in seq2seq models can be used for modeling confidence. \citep{knowles-koehn-2018-lightweight} points out the need for confidence scores for downstream uses of translation technology and attempts to derive them at the token (word) level. We wish to point out that making use of token level confidence scores poses difficulties because they cannot be treated independently within a given sentence.

A general study on ensemble methods for estimating uncertainty in structured prediction problems such as speech recognition and MT modeling is carried out in \citep{malinin2020uncertainty}; \citep{ott2018analyzing} and \citep{Xiao2019WatHJ} aim at modeling uncertainty in MT and provide valuable insights but do not go quite as far as estimating confidence.

\subsection{Related Work on Data Annotation}

As a departure from previous work on binary classification for QE, we set up our own data annotation framework approach as described in Section~\ref{sec:data}. We abandon the post-editing view altogether and simply ask annotators to label a translation as perfect ($good$) or not ($needs\ work$).

While cost-effective, crowd-sourced data is noisy as noted by \citep{turchi2014springer} in the context of QE in particular, and by \citep{TACL389} in a more general corpus generation context. We propose a simple model for annotation noise that allows us to estimate the probability of error and suggests modeling improvements such as use of label smoothing in training that proves fruitful.

\section{Data Annotation}
\label{sec:data}

Early experiments on WMT QE data and later ones on various internal data sets convinced us of the necessity to develop the data annotation capability targeting specifically the binary labeling $good\ /\ needs\ work$. Surrogate labeling derived by thresholding the edit distance between translation and reference translation, or any other measures of translation quality proved inadequate for our purposes: the correlation between such binary labels and those derived by thresholding the post-editing HTER score by labeling $good$ = (HTER == 0) was very poor.

We rely on annotations produced by two different annotator pools: one consisting of non-expert annotators proficient in the source and native in the target language, and a second one of expert annotators consisting of professional translators. A set of (source, target) sentence pairs, where the target is the translation of the source sentence as produced by a state-of-the-art MT model is partitioned into work items, each containing 6 sentence pairs. The source sentences are selected at random from web pages
sent as translation requests to Google Translate in the locale of interest by Google Chrome users. 

Each bilingual annotator is instructed as follows:
``
You will be asked to choose between two labels: 'good' and 'needs work' for the English translation of the input German sentence. The 'good' label should be assigned to perfect translations only.
If there are fluency problems in the English translation or context from the German sentence is required to restore the correct meaning for the English translation, please assign the 'needs work' label."
A few translations for the same input sentence are presented as examples, illustrating various translation problems and how they should be labeled.

We can request a variable number of ratings for each work item, each provided by a different annotator. The final label for a sentence pair is assigned by majority vote; to avoid ties, the number of ratings requested is an odd number. Initial experiments while developing the template requested five ratings per item and estimated the annotation error and confidence interval for the SACC measurements. English-Spanish CE development and test data (each comprising 1200 data samples) is annotated with 3 ratings per item/sentence pair whereas the 12,000 training data samples with just one rating per item/sentence pair to save costs. 

\begin{table}[ht]
    \centering
    \begin{tabular}{|l|c|c|c|}\hline
    Annotator Pool & Train   & Dev     & Test   \\
     (En-Es)   & SACC(1) & SACC(3) & SACC(3)\\\hline
    Non-expert & 0.825   & 0.890   & 0.887  \\
    Expert     & 0.602   & 0.625   & 0.590  \\\hline
    \end{tabular}
    \caption{English-Spanish training, development and test set sentence level accuracies (SACC).}
    \label{table:es_data}
    \vspace{0mm}
\end{table}

Table~\ref{table:es_data} shows the SACC(1) and SACC(3) values for the English-Spanish data used for building CE models on both expert and non-expert labeled data. The difference between SACC(3) and SACC(1), in particular on non-expert labeled data, prompted us to take a closer look at the quality of annotation.

As a measure of inter-annotator agreement we use Krippendorff's alpha (KA,~\citet{KrippendorffAlpha} as implemented in the NLTK toolkit, \url{https://github.com/nltk/nltk}) and also compute the labeling entropy per sample (E):
\begin{eqnarray}
E & = & 1/M \sum_{s=0}^{M-1} e_s \nonumber\\
e_s & = & - [p \cdot log(p) + (1 - p) \cdot log (1 - p)]\nonumber
\end{eqnarray} where $p = k(s)/5$; $M$ denotes the total number of samples and $k(s)$ is the count of $good$ labels for a given sample $s$. Values are reported in Table~\ref{tab:sacc5}\footnote{The minimum value of the entropy is $\min e_s = 0.0$, for $k = 0$ or $k = 5$, when all annotators agree; the maximum value of the entropy is $\max e_s = 0.97$, when only 2 out of 5 annotators agree.}.
\begin{table*}[ht]
    \centering
    \begin{tabular}{|l|c|c|c|c|c|c|c|}\hline
    \multicolumn{2}{|c|}{} & \multicolumn{3}{c|}{Non-expert} & \multicolumn{3}{c|}{Expert}\\
    Lang & N & SACC  & KA & E & SACC & KA & E\\\hline
    En-De    & 5 & 0.978 &  0.076 & 0.33        & 0.567 &  0.522 & 0.40 \\\hline
    En-Es    & 5 & 0.894 &  0.154 & 0.43        & 0.606 &  0.170 & 0.69 \\\hline
    En-Fr    & 5 & 0.894 &  0.221 & 0.34        & 0.439 &  0.297 & 0.59 \\\hline
    En-Ro    & 5 & 0.611 &  0.377 & 0.51        & 0.261 &  0.323 & 0.50 \\\hline
    \end{tabular}
    \caption{English-\{German, Spanish, French, Romanian\} Sentence Level Accuracy for N = 5 ratings per sample (SACC(5)), Krippendorff's alpha (KA) and annotation entropy (E) values.}
    \label{tab:sacc5}
    \vspace{0mm}
\end{table*}
Except for English-German, KA values indicate that the non-expert pool is only slightly less reliable than the expert pool. The generally low KA values\footnote{\url{https://w.wiki/eZ2} indicates that data sets with KA values below 0.667 are deemed unreliable.} confirm the accepted wisdom that the quality of translation for isolated sentences out-of-context is hard to evaluate.

\subsection{SACC Confidence Interval Estimation}
\label{confidence_interval}
For initial development of the annotator template and measuring SACC on ``in-domain" data we annotated 5-way a sample of 180 En sentences. Bootstrapping analysis assigns a confidence interval to the SACC(N) estimate for N values less than 5: a random trial is generated by sampling without replacement N (1 or 3) ratings out of the 5 for each sentence, after which we compute the SACC for the trial. The SACC mean and standard deviation are then computed across 100 such trials.
\begin{table*}[ht]
    \centering
    \begin{tabular}{|l|c|c|c|c|c|}\hline
    \multicolumn{2}{|c|}{} & \multicolumn{2}{c|}{Non-expert} & \multicolumn{2}{c|}{Expert}\\
    Lang & N & b-SACC  & 0.95 confidence    & b-SACC  & 0.95 confidence\\\hline
    En-De    & 3 & 0.950 & [0.948, 0.952] & 0.562 & [0.558, 0.565] \\
             & 1 & 0.882 & [0.878, 0.887] & 0.553 & [0.549, 0.556] \\\hline
    En-Es    & 3 & 0.874 & [0.871, 0.877] & 0.591 & [0.586, 0.595] \\
             & 1 & 0.821 & [0.816, 0.825] & 0.566 & [0.559, 0.573] \\\hline
    En-Fr    & 3 & 0.886 & [0.883, 0.889] & 0.459 & [0.455, 0.463] \\
             & 1 & 0.850 & [0.846, 0.854] & 0.480 & [0.474, 0.485] \\\hline
    En-Ro    & 3 & 0.600 & [0.596, 0.603] & 0.276 & [0.272, 0.280] \\
             & 1 & 0.600 & [0.595, 0.605] & 0.313 & [0.308, 0.317] \\\hline
    \end{tabular}
    \caption{Bootstrapping estimates for English-\{German, Spanish, French, Romanian\} Sentence Level Accuracy (b-SACC), along with confidence interval; for the non-expert pool b-SACC(3) and b-SACC(1) match empirical values but for the expert pool they do not.}
    \label{tab:bootstrap}
    \vspace{0mm}
\end{table*}

We were surprised by the large difference between SACC(1) and SACC(3) (as well as b-SACC(1) and b-SACC(3), respectively) on the non-expert annotator pool for German, Spanish and French. Since bootstrapping analysis predicts well the empirical SACC(1) and SACC(3) on the non-expert data, we propose a simple probability model that assumes labels for a given sample are independent. For the expert data this assumption does not hold. A better annotation model is clearly desirable, as shown by \citep{TACL389} in a different context. The next section describes the simple annotation model on non-expert ratings in more detail.

\subsection{Annotation Error Model for Non-expert annotator Data}
\label{annotation_error} 

To explain the large difference in SACC between N = 1 and N = 3 or 5 we propose a model where each annotation is the result of a stochastic labeling process: the annotator assigns the correct label with probability $1 - e$, or flips it with probability $e$, $1.0 \gg e$. This is a strong assumption, ratings of a given sample are expected to be correlated. We have found this model to operate correctly only on the data produced by non-expert annotator pool.

For a data set where the relative frequency of perfect translations is $f(good) = g$ and individual ratings are i.i.d. according to the Bernoulli error model above, we would be measuring $SACC(N)$ as a function of $N$ as follows: 
\begin{eqnarray}
    SACC(1) & = & g \cdot (1-e) + (1-g) \cdot e \nonumber\\
            & = & g - e(2g-1)\nonumber\\
    SACC(3) & = & g \cdot [(1-e)^3 + 3e(1-e)^2] + \nonumber\\
            &   & (1-g)\cdot[e^3 + 3e^2(1-e)] \nonumber\\
            & = & g\cdot(1-e)^2(1+2e) + \nonumber\\
            &   & (1-g)e^2(3-2e) \nonumber\\
            & \approx & g \cdot(1-e)^2(1+2e) \label{eq:pred_sacc3}
\end{eqnarray}
With $g = SACC(5)$ we have: $$e = \frac{SACC(5) - SACC(1)}{2 \cdot SACC(5)-1}$$ and we can check the extent to which our estimate for $SACC(3)$ matches the empirical one. 

Table~\ref{tab:estimated_error} shows the error probability $e$, empirical $SACC(3)$ values as well as values predicted by our annotation model. Despite its simplicity, it predicts $SACC(3)$ values surprisingly well: except for English-Romanian, the estimated values are well within the 0.95 confidence interval.
\begin{table}[ht]
    \centering
    \begin{tabular}{|l|c|c|c|}\hline
    Lang & Error Prob & \multicolumn{2}{c|}{$SACC(3)$}\\
             & $e$ & Bootstrap & Predicted \\\hline
      En-De   & 0.100 & 0.950     & 0.950 \\
      En-Es   & 0.093 & 0.874     & 0.872 \\
      En-Fr   & 0.056 & 0.886     & 0.886 \\
      En-Ro   & 0.050 & 0.600     & 0.607 \\\hline
    \end{tabular}
    \caption{Non-expert annotator pool: estimated annotation error probability $e$ and agreement between bootstrap estimate $b-SACC(3)$ and predicted values according to Eq.~(\ref{eq:pred_sacc3}).}
    \label{tab:estimated_error}
    \vspace{0mm}
\end{table}

Due to annotator availability constraints we choose to work on English-Spanish, see Section~\ref{sec:exps}. Since our training data is annotated by a single annotator (for both cost effectiveness and annotator availability reasons), we model the uncertainty on training labels by using label smoothing \citep{chorowski2016better} during training. The probability of error $e$ is one of the training hyper-parameters estimated on development data, see Section~\ref{sec:exps}.

Another observation is that the annotation model allows us to characterize human annotators performance in terms of Precision/Recall and compare that to the performance of our CE models:
\begin{eqnarray}
P & = & \frac{g \cdot (1 - e)}{g \cdot (1 - e) + (1 - g) \cdot e}\nonumber\\
R & = & 1 - e\nonumber
\end{eqnarray}

A human annotator for English Spanish operates at Precision 0.988 / Recall 0.907; for English German the operating point is Precision 0.998 / Recall 0.9. It also worth noting that as long as the MT model is fed ``in-domain" data a CE model for English-German would need to operate at a much higher Precision threshold: simply labeling all samples as $good$ translations would result in Precision = 0.978\footnote{This operating point would achieve higher Recall = 1.0 than a single human annotator; combined with the really high Precision value it brings into question the necessity of a CE model for high performing MT models.}, only slightly below single annotator performance.

\section{Modeling Approach}
\label{sec:modeling}

State-of-the-art models developed for MT QE \citep{wang-EtAl:2018:WMT4} usually have two components: a feature extractor (FE) and a predictor (P). In regular QE models, the predictor uses an output regression layer and is trained using either MAE or MSE loss. In our setup, P is a standard binary classification layer consisting of an affine transformation of a certain input dimensionality followed by soft-max trained using cross-entropy loss and label smoothing, as explained in Section~\ref{annotation_error}. 

To allow easy experimentation with various FEs, we abstract an API for FE that maps sequences of fixed dimensional features such as LM or MT encodings for the tokens in the source sentence, or MT decoder encodings or logits for the tokens in the target sentence, etc. to a fixed dimensionality representation. The reduction along the token position dimension is accomplished using a bi-directional LSTM followed by layer normalization~\citep{ba2016layer}. An arbitrary number of such FEs can be used in a CE by concatenating their output before feeding it to the P.

Although not a competitive approach to CE, we also experiment with the ``naive" model that uses as a single uni-dimensional feature the conditional log-probability assigned to the target sentence by the NMT model normalized by length. The predictor P computes the binary soft-max using a weight matrix and offset parameterized by two values each, for a total of 4 parameters.

\subsection{Features Derived from the Translation Model}
\label{sec:modeling:mt}

As baseline FE we use the input to the soft-max layer of the decoder of the NMT transformer model that produced the translations. This can be optionally augmented with the ``mismatch features" described in \cite{wang-EtAl:2018:WMT4} enriched with the per-token entropy of the decoder output probability distribution:
\begin{itemize}
    \setlength\topsep{-0.3em}
    \setlength\itemsep{-0.3em}
    \item is $\arg\max P(t_k)$ the same as the actual token $t_k^*$ at position $k$ in the target sentence? (binary)
    \item $\log(\arg\max P(t_k)) - \log(P(t_k^*))$
    \item $\log(\arg\max P(t_k)) + \log (Z)$ ``unnormalized log-probability", sometimes also referred to as ``logit") 
    \item $\log(P(t_k^*)) + \log (Z)$
    \item $H_k = \sum_{t_k} P(t_k) \log(P(t_k))$
\end{itemize}
where $t_k^*$ denotes the actual token at position $k$ in the target sentence whereas $t_k$ denotes a running token at position $k$.

As yet another FE derived from the underlying MT model we have the option of adding the source token encodings at the top layer. We have not used any features derived from the attention model but that is a promising extension for future work, in particular input coverage.

For CE model training back-propagation gradient is stopped at the underlying FE; dropout in the FE models is set to 0.0 to match it with inference.
All models are implemented in Lingvo \citep{shen2019lingvo}.

\subsection{Source Language Modeling Features}
\label{sec:modeling:lm}

LM-derived features on the source sentence are widely used in CE to be able to characterize the extent to which the input is ``in-domain". An LM is trained on the source side of the training data used for the MT model. Similar to the FE derived from the MT model, we use the LM embeddings (soft-max input) augmented with mismatch and entropy features. 

We extend the source-side LM FE by using a contrastive LM: a second (adapted) LM is trained on the CE training data incrementally from the previous one. The aim of this second LM is to capture differences between the domain in which we plan on using the MT and CE models and the domain in which the MT model was trained.

The same features as for the base LM are used for the adapted LM. In the case of using a contrastive LM FE, the concatenation of the feature sequences from the two LMs is augmented with two difference features and sent to the LSTM layer that encodes into a fixed dimensional feature:
\begin{itemize}
    \setlength\topsep{-0.3em}
    \setlength\itemsep{-0.3em}
    \item $\arg\max P_{base}(s_k) == \arg\max P_{adapted}(s_k)$ (binary)
    \item $\log P_{base}(s_k^*) - \log P_{adapted}(s_k^*)$
\end{itemize}

The perplexity difference between the base and adapted LMs on CE dev data is indicative of the mismatch between the domain on which the MT model was trained on and the domain on which we build and evaluate the CE model.

\subsection{Monte Carlo Dropout Features}
\label{sec:modeling:mcdropout}

In an attempt to characterize MT model uncertainty on a given input, we use Monte-Carlo dropout~\citep{gal2016dropout} for FE: we run the underlying MT model several times, each time with a different drop-out mask for the same drop-out probability value. The mean and variance for each of the target log-probabilities $\log P(t_k^*|s), \forall k = 0, \ldots, T-1$ are concatenated to the other target side MT-derived features described in Section~\ref{sec:modeling:mt} before LSTM encoding to fixed dimensionality.

\section{WWW Data Experimental Setup}
\label{sec:exps}

As described in Section~\ref{sec:data}, we target web page translation in a given locale.

Translation is done with a hybrid transformer-RNN encoder-decoder model~\citep{DBLP:journals/corr/abs-1804-09849} with multi-headed attention and shared word-piece vocabulary between source and target of size 32,000. The transformer encoder~\cite{vaswani2017attention} consists of 6 layers of dimension 1024, hidden dimension 8192 with residual connections. Decoder-encoder attention and encoder self-attention uses 16 heads. The decoder consists of 8 LSTM layers interleaved with layer normalization~\cite{ba2016layer}. The model has about 330 million parameters and is trained using dropout, label smoothing and AdaFactor optimization with a three stage schedule on a large amount of parallel data mined from the web (about 2 billion sentences).

For training the CE model we sample 14,400 sentences from web pages
sent as translation requests in the English-Spanish locale by Google Chrome users. 
12,000 are then annotated with $good\ /\ needs\ work$ labels by a single annotator using the annotation template described in Section~\ref{sec:data}. The remaining 2,400 sentences are annotated by three annotators each, the final label is assigned using majority voting and then split evenly between development and test data. 

The baseline CE model uses the MT model above to derive the features described in Section~\ref{sec:modeling:mt}. Training uses AdaFactor with a three stage schedule, dropout and exponential moving average decay as implemented in Lingvo~\citep{shen2019lingvo}. To match training with inference, dropout in the underlying MT model used for FE is set to 0.0 during training of the CE model; as mentioned in Section~\ref{sec:data}, label smoothing improves results significantly.

\subsection{Source LM Features}
\label{exps:lm_features}
As described in Section~\ref{sec:modeling:lm}, we trained two LSTM LMs (embedding dimension 1024, LSTM state dimension 2048, same vocabulary as the MT model):
\begin{itemize}
    \setlength\topsep{-0.3em}
    \setlength\itemsep{-0.3em}
    \item $LM_{base}$: trained on the source sentences in the MT model training data
    \item $LM_{adapt}$: trained on the source sentences in the CE training data, incrementally after initialization from $LM_base$
\end{itemize}

The perplexity of $LM_{base}$ measured on dev data matched to the training data is 41, compared to 53 when measured on the CE dev data. After incremental training for $LM_{adapt}$, the perplexities of $LM_{base}$ and $LM_{adapt}$ measured on the CE development data are very close as well: 53 and 51, respectively.
This shows that the CE data is ``in-domain" relative to the MT model training data; as expected, experiments using LM augmented CE models did not improve on the baseline.

\subsection{Monte Carlo Dropout}

We also conducted experiments that augmented the baseline FE with Monte-Carlo dropout features, as explained in Section~\ref{sec:modeling:mcdropout}. Since the computational overhead is significant, we tried a small number of trials (8, 16, 32) but observed no improvement over the baseline.

\subsection{Hyper-parameter Tuning}
For hyper-parameter tuning we use a black-box Bayesian optimization algorithm~\citep{frazier2018tutorial} exploring a grid:
\begin{itemize}
    \setlength\topsep{-0.3em}
    \setlength\itemsep{-0.3em}
    \item exponential moving average decay factor: [0.999, 0.9999] (0.999)
    \item number of LSTM layers: [2, 4, 8] (4)
    \item LSTM size: [32, 64, 128] (128)
    \item LSTM dropout probability: [0.3, 0.6] (0.6)
    \item learning rate: [1e-5, 5e-5, 1e-4] (5e-5)
    \item label smoothing probability: [0.07, 0.09, 0.11] (0.11)
    \item use nmt model encoder output: [true, false] (true)
    \item use meta features: [true, false] (true)
\end{itemize}

\subsection{Results}
Experimental results on the non-expert and expert data are presented in Tables~\ref{table:sq:results}-\ref{table:expert:results}. In both cases the naive model using just the conditional log-probability assigned by the NMT model to the target sentence performs poorly. For the non-expert data set there is a wide gap between the dev and test data performance, which brings into question the reliability of the Recall value on test data.

\begin{table*}[htbp]
    \centering
    \begin{tabular}{|l|c|c|c|}\hline
    Model                         & \multicolumn{3}{c|}{Recall at 0.95 Precision}\\
                                  & dev         &  test & training step \\\hline
    Naive (LogP)                  & 0.06        &  0.42  & 0.430k \\\hline
    Baseline                      & 0.50        &  0.57  & 1.740k \\
    LM on source                  & 0.52        &  0.45  & 0.716k \\
    Contrastive LM on source      & 0.57        &  0.45  & 0.572k \\\hline
    Monte Carlo dropout           & 0.59        &  0.43  & 0.476k \\
    LM on source and Monte Carlo dropout & 0.40 &  0.43  & 5.354k \\
    Contrastive LM on source and Monte Carlo dropout & 0.57  & 0.46 & 0.472k\\\hline
    \end{tabular}
    \caption{Recall at 0.95 Precision on non-expert data using various modeling approaches.}
    \label{table:sq:results}
    \vspace{0mm}
\end{table*}

For the non-expert data set, our best model uses the hyper-parameter values specified in parentheses above, and achieves 0.5 Recall at 0.95 Precision on dev data and 0.57 Recall on test data using a checkpoint picked on dev data after 1,740 training steps. We tried augmenting that baseline with either LM features on the source or Monte-Carlo dropout. The former did not help, as expected based on the perplexity measurements in Section~\ref{exps:lm_features}, and the latter did not help either. Exponential moving average decay is useful for regularization during training on such a small amount of data. Label smoothing is necessary to mitigate the error in training data labeling. Meta-features and encoder output are also useful.

For the expert data set we had to adjust the operating point of the CE model down to 0.9 Precision to be able to achieve significant Recall values. Even so, the highest Recall attained was 0.29; although noisy, the results suggest that Monte Carlo dropout and using a LM on the source could be marginally useful.

\label{exps:ewok_Expert_qe}
\begin{table*}[ht]
    \centering
    \begin{tabular}{|l|c|c|c|}\hline
    Model                         & \multicolumn{3}{c|}{Recall at 0.9 Precision}\\
                                  & dev         &  test & training step \\\hline
    Naive (LogP)                  & 0.01        &  0.08  & 0.091k \\\hline
    Baseline                      & 0.20        &  0.25  & 5.627k \\
    LM on source                  & 0.22        &  0.20  & 1.090k \\
    Contrastive LM on source      & 0.23        &  0.18  & 1.326k \\\hline
    Monte Carlo dropout           & 0.27        &  0.26  & 0.692k \\
    LM on source and Monte Carlo dropout & 0.26 &  0.29  & 1.022k \\
    Contrastive LM on source and Monte Carlo dropout & 0.19 & 0.24 & 1.450k \\\hline
    \end{tabular}
    \caption{Recall at 0.9 Precision on expert data using various modeling approaches.}
    \label{table:expert:results}
    \vspace{0mm}
\end{table*}

\section{Conclusions and Future Work}
\label{sec:conclusions}

We have investigated the feasibility of confidence estimation for neural machine translation models operating at the high end of the performance spectrum. As a side product of the data annotation process necessary for building such models we propose sentence level accuracy $SACC$ as a simple, self-explanatory evaluation metric for translation models. Experiments on two different annotator pools, one comprised of non-expert (crowd-sourced) and one of expert (professional) translators show that $SACC$ can vary greatly depending on the translation proficiency of the annotators, despite the fact that both pools are about equally reliable according to Krippendorff's alpha (KA) metric. The relatively low values of KA confirm the expectation that binary quality labeling $good$ / $needs\ work$ for translation of sentences out of context is very hard.

We have shown that for an English-Spanish translation model operating at $SACC = 0.89$ according to a non-expert annotator pool we can derive a confidence estimate that labels 0.5-0.6 of the $good$ translations in an ``in-domain" test set with 0.95 Precision. Using a simple probabilistic annotation model we estimate the error probability for non-expert labeling and find it to be 0.09. The annotation error model allows us to estimate the Precision/Recall ``operating point" for a single human annotator: for English-Spanish it is Precision 0.988 / Recall 0.907, significantly better than our CE model and leaving plenty of room for improvement. For English-German the single annotator operating point is Precision 0.998 / Recall 0.9, significantly raising the bar for a CE model: simply labeling all samples as $good$ translations would result in Precision = 0.978 / Recall 1.0, only slightly below single annotator performance and bringing into question the need for CE for such high quality MT models.

Switching to an expert annotator pool decreases $SACC$ dramatically: $0.61$ for English-Spanish and $0.57$ for English-German, measured on the exact same data as above. This forces us to lower the CE model operating point to 0.9 Precision while labeling correctly about 0.25 of the $good$ translations in the data. The simplistic error model proposed for the non-expert ratings does not fit the data well in this case, leaving ample room for further research on this particular topic.

As a broader conclusion, confidence estimation is not just a function of the translation model that it operates on (and implicitly the training data used to build it) but depends also on two other important factors: the data that the translation model is used on and for which the confidence score is desired, as well as the level of proficiency of the annotator pool used for labeling the data necessary for building the confidence estimation model. It is critical that the latter matches the end-user expectation for translation quality in the desired domain. 

For future work we would like to reduce the model's dependency on labeled training data with the twin goals of ease of deployment and improved portability across domains. Fully unsupervised (no training data) CE modeling would enable predicting SACC on a test set, offering an alternative to~\citet{fomicheva2020unsupervised}.

We would also like to augment the model with features derived from N-best lists or posterior probability lattices, attention matrix as in \citep{DBLP:journals/corr/abs-1710-03743}, and study the model behavior and utility on out-of-domain data as well as for languages where translation accuracy is lower than for English-Spanish. 

Aggregating sentence level confidence scores for larger units of text (paragraphs or documents) is also a promising direction for improving the reliability and usefulness of the confidence estimate. 

\section{Acknowledgments}

We would like to thank George Foster for comments and suggestions, Jonathan Shen for help with implementation design and Lingvo support, Bowen Liang and Klaus Macherey for assistance with models trained on parallel data mined from the web. 


\bibliographystyle{acl_natbib}
\bibliography{emnlp2020}



\end{document}